\documentclass[journal]{IEEEtran}

\ifCLASSINFOpdf
\else
\fi
\usepackage{cite}
\usepackage{amsmath,amssymb,amsfonts}
\usepackage{algorithmic}
\usepackage{algorithm}
\usepackage{graphicx}
\usepackage{subfigure}
\usepackage{textcomp}
\usepackage{booktabs}
\usepackage{graphicx}
\usepackage{amsmath}
\usepackage{subfigure} 
\usepackage{tabularx}
\usepackage[table,xcdraw]{xcolor}
\usepackage[colorlinks,linkcolor=blue,citecolor=blue]{hyperref}
\usepackage{svg}
\def\BibTeX{{\rm B\kern-.05em{\sc i\kern-.025em b}\kern-.08em
    T\kern-.1667em\lower.7ex\hbox{E}\kern-.125emX}}

\begin{document}

\title{Continual Adaptation for Autonomous Driving \\ with the Mixture of Progressive Experts Network}

\author{Yixin Cui, Shuo Yang, Chi Wan, Xincheng Li, Jiaming Xing,\\
Yuanjian Zhang, Yanjun Huang, and Hong Chen,~\IEEEmembership{Fellow,~IEEE}

\thanks{Manuscript received 20 January 2025. 
This work was supported by the National Key Research and Development Program of China under Grant 2022YFB2502900 and the National Natural Science Foundation of China, Joint Fund for Innovative Enterprise Development (U23B2061). (Corresponding author: Yanjun Huang.)}
\thanks{Yixin Cui, Shuo Yang, Xincheng Li, Jiaming Xing and Yuanjian Zhang are with the School of Automotive Studies, Tongji University, Shanghai 201804, China (e-mail: 2411448@tongji.edu.cn; 2111550@tongji.edu.cn; 2111302@tongji.edu.cn; 2210809@tongji.edu.cn; 24093@tongji.edu.cn).}
\thanks{Chi Wan is with the College of Mechanical and Vehicle Engineering, Chongqing University, Chongqing 130012, China (e-mail: 20212243@stu.cqu.edu.cn).}
\thanks{Yanjun Huang is with the School of Automotive Studies, Tongji University, Shanghai 201804, China, and also with the Frontiers Science Center for Intelligent Autonomous Systems, Shanghai 200120, China (e-mail: yanjun\_huang@tongji.edu.cn).}
\thanks{Hong Chen is with the College of Electronics and Information Engineering and the Clean Energy Automotive Engineering Center, Tongji University, Shanghai 201804, China (e-mail: chenhong2019@tongji.edu.cn).}
}

\markboth{IEEE Transactions on Intelligent Transportation Systems}%
{Shell \MakeLowercase{\textit{et al.}}: Bare Demo of IEEEtran.cls for IEEE Journals}

\maketitle

\begin{abstract}
Learning-based autonomous driving requires continuous integration of diverse knowledge in complex traffic , yet existing methods exhibit significant limitations in adaptive capabilities. Addressing this gap demands autonomous driving systems that enable continual adaptation through dynamic adjustments to evolving environmental interactions. This underscores the necessity for enhanced continual learning capabilities to improve system adaptability. To address these challenges, the paper introduces a dynamic progressive optimization framework that facilitates adaptation to variations in dynamic environments, achieved by integrating reinforcement learning and supervised learning for data aggregation. Building on this framework, we propose the Mixture of Progressive Experts (MoPE) network. The proposed method selectively activates multiple expert models based on the distinct characteristics of each task and progressively refines the network architecture to facilitate adaptation to new tasks. Simulation results show that the MoPE model outperforms behavior cloning methods, achieving up to a 7.8\% performance improvement in intricate urban road environments.
\end{abstract}


\IEEEpeerreviewmaketitle


\section{Introduction}
\IEEEPARstart{A}{utonomous} driving technology is a pivotal force in the evolution of future transportation, with profound  significance in enhancing road safety and advancing intelligent mobility \cite{ahmed2022technology} \cite{huang2023efficient}. Learning-based algorithms, such as deep learning and reinforcement learning, exhibit substantial potential in this field by significantly enhancing the intelligence of autonomous driving systems through learning continuously from extensive data \cite{ly2020learning,feng2023dense,riemer2018learning}. Their adaptive and evolving capabilities make these algorithms indispensable for achieving high-level autonomous driving, thus laying a solid technical foundation for intelligent transportation systems \cite{khilenko2018solving}.
Among various learning-based algorithms, imitation learning methods have emerged as a crucial research direction for end-to-end autonomous driving, primarily due to their capability to effectively simulate human driving behaviors \cite{pan2020imitation}. Behavior cloning, a key imitation learning technique, directly maps sensory inputs (e.g., camera images, state vectors) to vehicle controls (e.g., steering, acceleration, braking) by minimizing the difference between model outputs and human driving actions through parameter optimization \cite{wang2022high}. The end-to-end design of behavior cloning not only significantly reduces system complexity but also exhibits strong robustness across diverse tasks \cite{chen2024end}. Based on fundamental tasks such as route following, researchers have developed various imitation learning methods to tackle challenges posed by more complex driving scenarios \cite{ly2020learning} \cite{li2022driver}.

However, the long-tailed distribution problem represents a critical challenge in autonomous driving, significantly constraining the performance of imitation learning \cite{chen2024end}. The fundamental challenge stems from the stark contrast between common routine scenarios and rare yet diverse safety-critical scenarios. For instance, complex intersection navigation and emergency collision avoidance, though occurring infrequently, are crucial for ensuring driving safety and system robustness \cite{zhou2022dynamically}. The long-tailed distribution inherent in imbalanced data limits imitation learning's adaptability, ultimately compromising system robustness and safety assurance . As a fundamental requirement for autonomous systems, adaptability remains a major hurdle for current imitation learning methods, particularly in managing complex, dynamic environments \cite{peng2021long}.

In recent years, self-learning algorithms have emerged as a promising approach, demonstrating remarkable adaptability through environmental interaction and dynamic strategy optimization \cite{xing2024comprehensive,kiran2021deep,zhao2024survey}. A key strength lies in their reduced dependence on large-scale datasets, which helps alleviate the long-tailed distribution challenge and enhances generalization to new scenarios \cite{chiba2021effectiveness} \cite{kirkpatrick2017overcoming}. By leveraging advanced self-learning mechanisms such as transfer learning, reinforcement learning, and meta-learning, these algorithms exhibit efficient knowledge transfer and rapid task adaptation, even under data-constrained conditions \cite{feng2023dense} \cite{sallab2017meta}. While offering some distinct advantages over imitation learning, these algorithms still face notable limitations in their adaptive capabilities. Self-learning algorithms continue to struggle with dynamic and unpredictable autonomous driving applications. Their limited cross-domain adaptability manifests in delayed strategy adjustments during complex task transitions, compromising system robustness against environmental uncertainties \cite{huang2023efficient}.

To address these limitations, we propose a new continuous adaptation mechanism that synergistically optimizes data processing and model architecture for enhanced autonomous driving adaptability. Our methodology establishes an integrated learning framework that harmoniously combines self-learning and imitation learning approaches, providing an effective solution to long-tailed distribution issues via adaptive model refinement. This dynamic architecture facilitates continuous performance refinement during operation, significantly boosting decision-making capabilities in complex environments. Furthermore, we propose a multi-expert network that intelligently synthesizes domain expertise through dynamic network expansion. This innovative component substantially improves adaptability and decision robustness of the system. The experimental results substantiate that our framework achieves better performance than existing approaches, offering a robust, scalable, and continuously adaptive solution for autonomous driving systems operating in complex scenarios.

The contributions of this study are summarized as follows:
\begin{enumerate}
    \item A dynamic progressive optimization framework is introduced, incorporating a data aggregation approach that combines reinforcement and supervised learning for rapid adaptation to dynamic environments and accelerate model adaptation with generated high-quality data.
    
    \item As a key component of this framework, the Mixture of Progressive Experts (MoPE) model is proposed. It enables the adaptation to new tasks while preserving knowledge of previous tasks through a expandable hybrid expert paradigm.
    
    \item The gating mechanism dynamically adjusts expert weights, prioritizing experts that match the task and integrating multi-expert knowledge, thereby significantly enhancing the system's adaptability in diverse scenarios.
\end{enumerate}

This paper is organized as follows. The architecture of the dynamic progressive optimization framework is introduced in Section II. The Mixture of Progressive Experts method is formulated in Section III. In Section IV, the proposed method is validated and analyzed under various scenarios, and Section V concludes this paper.

\section{Proposed Framework}
In autonomous driving, while efficient algorithms are pivotal for model optimization, high-quality and sufficiently large expert data is equally indispensable, particularly in dynamic and unpredictable environments\cite{liu2024survey}. Achieving enhanced model adaptability requires comprehensive, heterogeneous, and high-quality datasets for both training and evaluation. To bridge the gap between algorithmic efficiency and data dependency, it is essential to establish a well-structured robust data processing pipeline. Such a pipeline must integrate efficient data generation, continuous data input, and adaptable model architectures to enable continuous adaptation. Additionally, it ensures that data flow aligns seamlessly with algorithmic demands.
\begin{figure*}[htbp]
\centering
\includegraphics[width=6.4in]{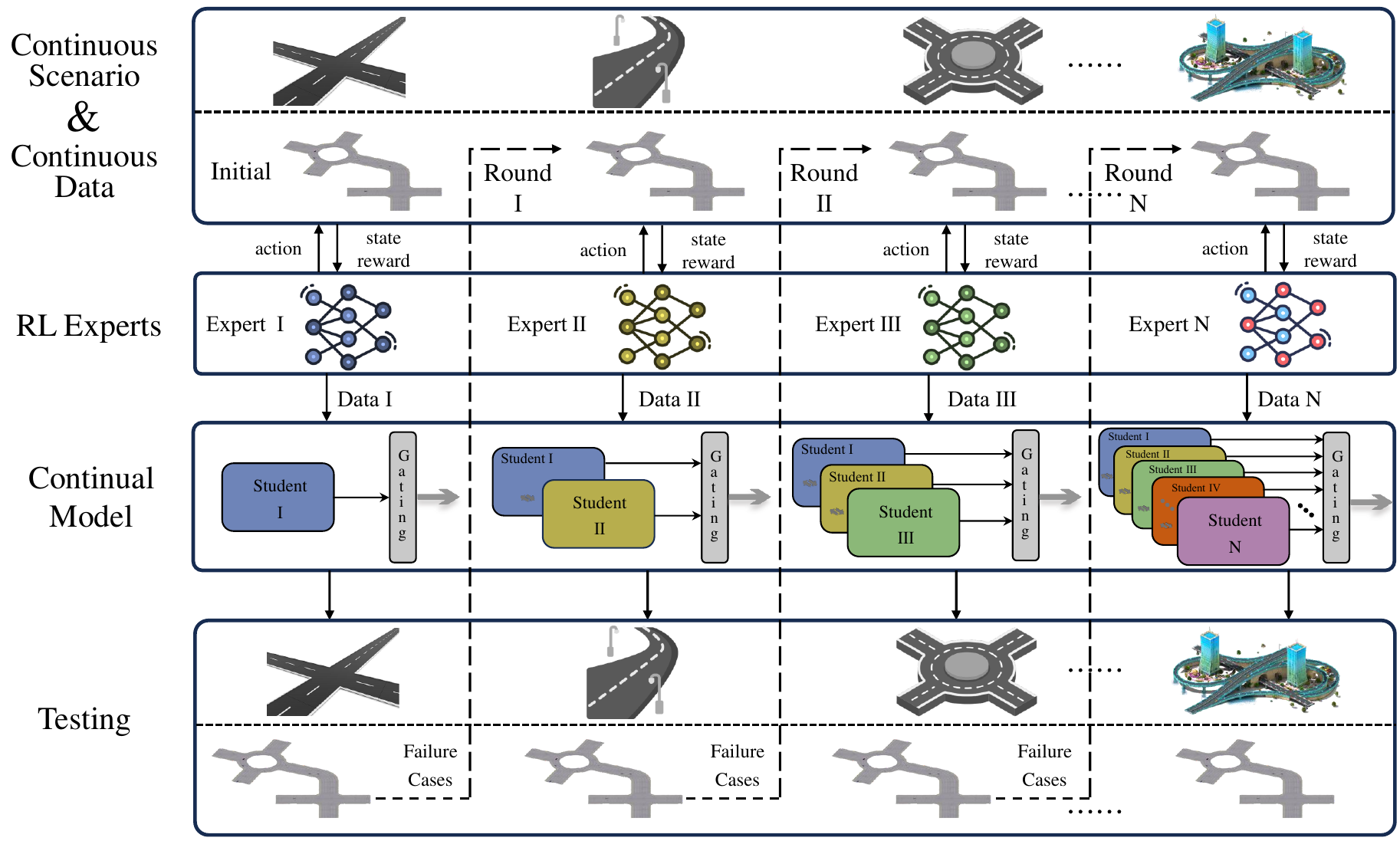}
\caption{Overview of dynamic progressive optimization framework for autonomous driving. Reinforcement learning experts generate high-quality, multi-scenario data in simulation, which is then aggregated by the Continual Model using imitation learning. The model progressively expands and dynamically weights expert modules to adapt to new scenarios or cases while retaining past knowledge, ensuring robust performance in complex traffic environments.
}

\label{fig1} 
\end{figure*}

In this paper, we introduce the dynamic progressive optimization framework, designed to address the challenges of continuous adaptation in dynamic environments. As depicted in the Fig.\ref{fig1}, the framework facilitates the progressive optimization of expert agents through data aggregation. Furthermore, it supports continuous adaptation via a multi-expert strategy distillation model, termed the Continual Model $\mathcal{N}$, which dynamically integrates knowledge from multiple expert agents to adapt to evolving tasks \cite{hong2024knowledge}.

As the initial phase of the framework, reinforcement learning is employed to generate robust and varied datasets across multiple scenarios, accumulating driving strategy knowledge through trial-and-error processes in a simulated environment \cite{zhang2021end}. Subsequently, the Continual Model incrementally trains the data using an imitation learning approach, progressively integrating new scenarios and datasets. Subsequently, the Continual Model adopts a continual imitation learning approach to incrementally incorporate new scenarios and datasets, facilitating progressive task adaptation \cite{wan2024lotus}. Through data aggregation, decision-making insights from multiple experts across various tasks are consolidated, while policy distillation extracts and transfers critical knowledge to optimize performance. Ultimately, this dynamic progressive optimization framework, embodied in the Continual Model, develops an expert agent capable of making efficient, real-time decisions in complex and dynamic environments, manifesting exceptional long-term adaptability. The subsequent part provides a detailed overview of the specific technical implementation of the dynamic progressive optimization framework.

In the implementation of the dynamic progressive optimization framework, multiple expert agents are initially trained using reinforcement learning algorithms. Each expert executes distinct policies and generates data across a variety of training settings. The reinforcement learning process for autonomous driving can be viewed as a Markov Decision Process $\mathcal{M} = (S; A; P; R)$, where $S$ is the state space, $A$ is the action space, $P$ is the transition model, and $R$ is the reward function. In this process, the agent interacts with the environment and takes actions based on the current state to achieve optimal performance, as described by the following equation:
\begin{equation}
\begin{aligned}
\pi(a|s) = P(A = a | S = s)
\end{aligned}
\end{equation}
where the policy function $\pi(a|s)$ represents the probability of taking action $a$, when given state $s$. In this paper, we adopt a state-of-the-art method in reinforcement learning, the Soft Actor-Critic (SAC) algorithm \cite{haarnoja2018soft}. The optimal policy $\pi^*$ is computed as:
\begin{equation}
\begin{aligned}
\pi^* = \arg \max_{\pi} \mathbb{E}_{(s, a) \sim \rho_\pi} \left[ \sum_t r(s, a) + \lambda \cdot \mathcal{H}(\pi(\cdot | s)) \right]
\end{aligned}
\end{equation}
where the objective combines the reward function $r(s, a)$ with the entropy term $\mathcal{H}(\pi(\cdot|s))$ , balancing reward optimization and exploration through the regularization parameter $\lambda$.

The state space for the autonomous vehicle is normalized and comprises three key elements:
\begin{equation}
\begin{aligned}
s=[S_{ego},S_{navi},S_{lidar}] 
\end{aligned}
\end{equation}
where the ego state  $S_{ego}$  provides the agent with information about the vehicle itself, the navigation state $S_{navi}$ supplies details about the navigation map, and the environment state $S_{lidar}$ represents surrounding obstacles using processed LiDAR point cloud data arranged clockwise from the vehicle's front. The agent acts based on the current state, where the action space includes steering, acceleration, and brake signals. Specifically, the action is normalized as a vector $a =[u_s , u_a]$, where $u_s$ is the steering signal, $u_a$ represents the acceleration and brake signal. The reward function $R$ consists of dense driving reward, speed reward, and sparse termination reward. The driving reward $R_d$ encourages movement along the reference lane and estimates the acceleration for the speed reward $R_s$, while the termination reward $R_t$ is added at the end of the episode, such as when reaching the destination or encountering a collision. This can be expressed mathematically as:
\begin{equation}
\begin{aligned}
R &= c_1 R_d + c_2 R_s + R_t \\
  &= c_1 (d_t - d_{t-1}) + c_2 \frac{v_t}{v_{\text{max}}} + R_t
\end{aligned}
\end{equation}
where $d_t$ and $d_{t-1}$ represent the longitudinal position difference of the target vehicle at consecutive time steps.$v_t$ is the current velocity, and $v_{\text{max}}$ is the maximum velocity. $c_1$, $c_2$ are coefficients.

The outputs are continuous driving actions, including acceleration, deceleration, and steering. These reinforcement learning (RL) experts undergo extensive trial-and-error learning within the simulated environment. They are given the current state in each scenario, make decisions based on their policies, receive corresponding rewards, and iteratively refine their policies, thereby optimizing their driving behaviors over time. After repeated interactions, each RL expert masters the relatively optimal driving strategy for various scenarios, accumulating high-quality datasets (e.g., Data I, Data II, Data III). These datasets include the state-action pairs and action distributions generated by the experts. Since the data is generated in a simulated environment with minimal human intervention, it can be dynamically produced at scale, significantly reducing the costs and challenges of real-world data acquisition.

Through continuous interaction between deep neural networks and the traffic environment, autonomous driving strategies evolve over time, progressively achieving the optimal mapping from environmental states to executed actions. Furthermore, the high-quality datasets reflect the decision-making processes of expert agents across diverse driving scenarios, highlighting the variability in data distribution. By minimizing the following loss function, the model's predicted actions are optimized to closely match the expert's actions:
\begin{equation}
\begin{aligned}
L(\theta) = \mathbb{E}_{(s, a_e) \sim D} \left[ \left\| \mu(s; \theta) - a_e \right\|^2 \right]
\end{aligned}
\end{equation}
where $\mu(s; \theta)$ is the policy of the learned network model, and $a_c$ is the expert's action label.

The Continual Model $\mathcal{N}$ learns through imitation from high-quality data generated by expert agents, performing rapid adaptation through supervised learning. The input data format for the continual model aligns with the state representation used in reinforcement learning. This significantly improves both training efficiency and model performance. During the training process, the model iteratively updates its parameters by imitating the actions taken by the expert in each state. It minimizes the error between the model’s output and the expert’s action labels, effectively replicating the expert’s driving behaviors. The model’s goal is to capture the underlying strategy behind the demonstrated state-action pairs. The detailed explanation of the Continual Model construction is presented in Section \ref{sec:methodology}.

The Continual Model is trained in phases, with each phase introducing a new sub-expert model (Expert I, Expert II, etc.). Composed of these sub-experts, the model learns from expert data generated by RL agents, adapting to new scenarios while integrating past knowledge. The dynamic progressive optimization framework ensures the model continuously enhances its expertise over time, advancing performance to accommodate evolving demands.

The dynamic progressive optimization framework optimizes adaptability by enhancing both data and algorithmic elements. In terms of data, it enables the Continual Model to consistently learn from scenario-specific and safety-critical data, ensuring robust data integration. In the realm of algorithms, the Continual Model $\mathcal{N}$ is implemented through the proposed Mixture of Progressive Experts (MoPE). The detailed formulation and architecture of the model will be thoroughly explored in Chapter III. Addressing these issues is vital for creating a harmonious interplay between data and algorithms, ultimately driving advancements in autonomous driving adaptation. 


\section{The Mixture of Progressive Experts Network} \label{sec:methodology}

\subsection{Problem formulation}

To address the challenge of adaptability in continual learning settings, we first formalize the problem outline. In non-stationary environments characterized by evolving non-IID task distributions, the primary adaptability challenge stems from maintaining consistent performance across both new and previous tasks. It originates from neural networks' optimization bias, their tendency to favor current task performance over retaining prior knowledge, termed as catastrophic forgetting. Thus, the core objective is to develop a model capable of simultaneously acquiring new knowledge while retaining proficiency across all learned data.

We formulate the problem as sequential adaptation to data streams $\mathcal{W} = \{w_1, \dots, w_n\}$. In autonomous driving, each task $w_i$ corresponds to a specific driving task, such as handling intersection navigation or driving through urban traffic. Each one is associated with a distinct data distribution $P_i(X, Y)$. The inputs $X$ to the continual model consist of state vectors, which contain information about the self-vehicle, navigation, and surrounding environment, totaling 259 dimensions. This input vector is represented as:
\begin{equation}
\begin{aligned}
X = & \ [S_{\text{ego}}^1, S_{\text{ego}}^2, \ldots, S_{\text{ego}}^m, S_{\text{navi}}^1, S_{\text{navi}}^2, \ldots, S_{\text{navi}}^m, \\
& \ S_{\text{lidar}}^1, S_{\text{lidar}}^2, \ldots, S_{\text{lidar}}^o]
\end{aligned}
\end{equation}
with values $m = 9, \, n = 10, \, o = 240$. Here, $Y$ represents the vehicle's control actions (e.g., steering, acceleration, and braking).

Each distribution can be abstracted as a hyperparameterized linear regression problem, formally defined as \cite{li2024theory}:
\begin{align}
h(X) = X^\top w
\end{align}
where $w$ is the true parameter vector. This linear regression formulation provides an interpretable framework for modeling task-specific relationships through linear mappings, offering both computational simplicity and structural clarity \cite{chen2022towards}. 

In the over-parameterized case, multiple linear models can fit the data perfectly. The discrepancy between the optimal parameters for any two tasks satisfies:$\|w_i - w_j\|_{\infty} = O(\sigma_0)$. The optimization target is to minimize the total loss across all observed distributions:
\begin{align}
\mathcal{L} = \sum_{i=1}^N \mathbb{E}_{(X, Y) \sim P_i}[\ell(f(X; \theta), Y)]
\end{align}

However, as learning progresses, the growing heterogeneity of data distributions induces conflicting optimization dynamics. Specifically, given two disparate data distributions $w_i$ and $w_j$, with corresponding loss functions $\mathcal{L}_i$ and $\mathcal{L}_j$, if the respective parameter gradients $g_i=\nabla_\theta \mathcal{L}_i$ and $g_j = \nabla_\theta \mathcal{L}_j$ are in opposite directions: $g_i \cdot g_j < 0$. It indicates a conflict in the parameter updates required for these tasks. Such conflicts accumulate as more data is introduced, ultimately compromising the model's adaptability. Given the optimal parameters $\theta_i^*$ for each task, the lower bound of total loss can be expressed as:
\begin{align}
\mathcal{L} \geq \sum_{i=1}^N \|\theta - \theta_i^*\|^2
\end{align}
where $\|\theta - \theta_i^*\|^2$ denotes the deviation between the model’s current parameter $\theta$ and the optimal parameter $\theta_i^*$ for each task. As the number of tasks $N$ increases, the model must simultaneously approximate multiple optimal parameters $\theta_i^*$, which become highly challenging in high-dimensional parameter spaces. This challenge is particularly pronounced when there is little overlap between task parameters, as conflicts between the optimal solutions for different tasks become more significant, leading to a marked decline in adaptability.

To address this issue, we propose a model growth approach that enables the model to expand its capacity as new tasks are introduced. When new data is introduced, the model grows by adding dedicated parameters or components specific to the current task while retaining the parameters learned from previous tasks. This approach effectively reduces conflicts between tasks, ensuring that the model maintains performance across all tasks and remains adaptable as the number of tasks increases. Through this strategy, the model can efficiently allocate new parameters, progressively approximating the optimal parameters $\theta_i^*$ for each task. This model growth strategy ensures that the model can handle multiple tasks in high-dimensional parameter spaces while maintaining superior adaptability.


\subsection{Proposed model design}

Displayed in Fig.\ref{fig2} below, the proposed Mixture of Progressive Experts (MoPE) model addresses the challenge of efficient multi-expert load balance in a continual learning setting by progressively expanding the network architecture. By employing a gating mechanism to sparsely allocate tasks among experts and dynamically optimizing their weights, this approach effectively alleviates the risk of catastrophic forgetting. This approach effectively mitigates catastrophic forgetting by enabling the reuse of historical knowledge, while ensuring progressive improvement in model performance. Each time a new task is introduced, MoPE constructs an independent task-specific branch network and freezes the parameters of existing branches to preserve previously learned knowledge. New tasks are processed through the newly added branch networks, while lateral connection mechanisms allow for the sharing of historical information, enabling dynamic transfer and accumulation of knowledge \cite{liu2018progressive}. This design facilitates the learning of new tasks without overwriting prior knowledge, thereby providing strong support for task adaptability and knowledge robustness in continual learning scenarios. This section further delves into the theoretical foundation of the Mixture of Progressive Experts model from the continual learning perspective, emphasizing mechanisms that enhance adaptability in data stream environments \cite{vats2024evolution}.

Specifically, each branch network of the Mixture of Progressive Experts functions as an expert, with the model comprising $M$ experts in total. Each expert $m$ is characterized by its model parameters $w_t^{(m)}$ and a corresponding gating network $g_t^{(m)}$. Each task branch in the model has $L$ layers, focusing on a specific task during the $t$-th round of training. During model initialization, the parameters of new experts are set to zero, i.e., $\theta_0^{(m)} = 0$ and $w_0^{(m)} = 0$. Task assignment and parameter updates are executed in the first training round, with expert parameters being dynamically adjusted in subsequent rounds as new tasks arrive.
\begin{figure}[htbp]
\centering
\includegraphics[width=3.2in]{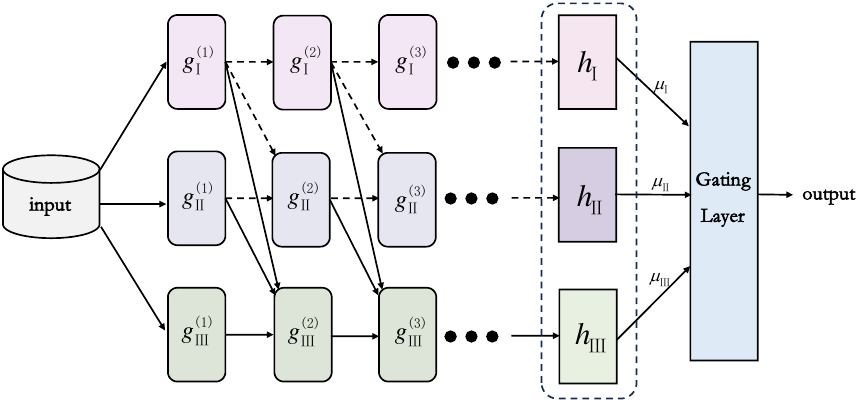}
\caption{The diagram of the Mixture of Progressive Experts.}
\label{fig2} 
\end{figure}

The continuous input training data consists of $N$ rounds, denoted as the dataset $W = \{w_1, \ldots, w_n\}$. As outlined in Algorithm 1, the Mixture of Progressive Experts training process begins by initializing the first branch network $\Phi^{(0)}$ and its gating network $g^{(0)}$ using the first round of task data $W_0$. During this training process, an appropriate learning rate $\eta$ and regularization coefficient $\lambda$ are set to control the model's learning speed and prevent overfitting. After the training of the first branch network in this stage is completed, the initial parameters of the model are obtained for subsequent task learning, and the weights are frozen. In each subsequent round of training $i \in \{ 1,2, \dots, N\}$, the model is updated iteratively by loading the current task data $W_i$ along with a portion of historical data $\overline{W}_{0:i-1}$, which are combined to form a new training dataset $\widetilde{W_i}$. A new $i+1$-th branch network $\Phi^{(i)}$ is added to the model, and a new gating network $g^{(i)}$ is assigned to this branch network. The new branch network focuses on feature learning for the current task. Each task corresponds to a branch network, allowing the model to gradually accumulate knowledge of different tasks while preventing the weights of old tasks from being overwritten. During the initial branch training, only the data from the first task is used, and its parameters are frozen. Subsequently, whenever a new task arrives, a new branch is instantiated and forms a lateral connection with the previous branch. Through this lateral connection, the new task can leverage the knowledge of existing tasks.

\begin{algorithm}
\caption{The Mixture of Progressive Experts Training}
\begin{algorithmic}[1]
\REQUIRE Training expert data $W = \{w_1, \ldots, w_N\}$
\REQUIRE Initial branch index $K$ (initially $K = 1$)
\REQUIRE Learning rate $\eta$, regularization factor $\lambda$, tolerance value $\Gamma$
\STATE Initialize the first-branch network $\Phi^{(0)}$ of the Continual Model $\mathcal{N}$, train using $W_0$, and set parameters $\theta^{(0)}$
\STATE Train $\theta^{(0)}$ until convergence and parameter freezing
\FOR{$i = 1$ to $N$}
    \STATE Load the data from the current iteration $W_i$, along with a subset of historical data $W_{0:i-1}$, to construct the final combined training dataset $\tilde{W}_i$
    \STATE Add a new branch network $\Phi^{(i)}$ to the model, and set the gating network parameters $\omega^{(i)}$
    \STATE Train the $(i+1)$-th branch network $\Phi^{(i)}$ and the corresponding gating network $g_i$ using $\tilde{W}_i$
    \IF{Output difference $\delta >$ tolerance value $\Gamma$}
        \STATE Update $\theta^{(i)}$ using gradient descent
        \STATE Recalculate the output difference
    \ENDIF
    \STATE Freeze and finalize the parameters of the $(i+1)$-th branch network $\theta^{(i)}$
\ENDFOR
\RETURN the Continual Model $\mathcal{N}$
\end{algorithmic}
\end{algorithm}

The Mixture of Progressive Experts network achieves knowledge sharing and efficient learning between tasks through a lateral connection mechanism, progressively expanding the network structure. For each new task, the branch network not only utilizes its own parameters but also dynamically extracts knowledge from historical tasks through lateral connection parameters $U_i^{(l)}$, while maintaining independent task optimization paths. This approach avoids overwriting existing knowledge and enhances adaptability between tasks. In each training round, the current round's branch network collaborates with historical branches through cross-branch connections, enabling knowledge sharing and feature expansion. Each branch network consists of a multi-layer modular structure, where each module captures the core features of the current task while integrating knowledge from historical tasks through lateral connections. Features are accumulated in a layer-by-layer recursive manner, and outputs are optimized using nonlinear activation functions. This design allows the model to efficiently adapt to new tasks in a continual learning environment while retaining memories of old tasks, enhancing the transfer and adaptability of cross-domain knowledge. The output of each layer follows the principles of recursive weighting and feature aggregation.

Assume the input for the $t$-th task is $x_t$, and the model constructs a new branch network for the $t$-th task. The output $h_t^{(l)}$ of the $l$-th layer in each branch network depends on the input of the current task and the outputs of historical branches. In the MoPE model, the output of each module adheres to a recursive computation principle. Specifically, for all layers except the last layer, the output is calculated as follows. In this formulation, $y_t^{(l)}$ represents the output of the $l$-th layer at the $t$-th round, computed as a ReLU activation over both the current input and historical information:
\begin{align}
y_t^{(l)} = \text{ReLU}\left( V_t^{(l)} x_t^{(l)} + \sum_{i=1}^{t-1} U_i^{(l)} h_i^{(l)} \right)
\end{align}
where $V_t^{(l)}$ is the linear layer parameter for the $l$-th layer in the current round, acting on the input $x_t^{(l)}$, which is the output from the preceding layer. The terms $U_i^{(l)}$ capture lateral connections from the $i$-th historical branch, using $h_i^{(l)}$, the output of the $l$-th layer of that branch. By combining these components and applying the ReLU function, the model integrates both the current round's input and learned representations from previous branches.

This formula reflects the mechanism of knowledge sharing between tasks, where the output of the current task depends not only on its own input but also integrates features from historical tasks through auxiliary connection parameters $U_i^{(l)}$. Through lateral connections, the model achieves dynamic aggregation and transfer of knowledge in a layer-by-layer accumulation process. In the last layer, the model directly generates the output, following the same computation logic as the previous layers but removing the activation function to achieve the final weighted combination of features or loss optimization. The feature learning in each layer originates from both the input of the current task and the knowledge of historical tasks effectively introduced through the lateral connection mechanism, enabling knowledge reuse and fusion. This layer-by-layer recursive structural design ensures that the model can efficiently inherit and extend knowledge from historical tasks even with frozen parameters, enhancing its cross-domain adaptability.

The model consists of $M$ experts, each represented by a set of parameters $\theta^m$. To optimize adaptability, the gating network dynamically assigns weights $\Theta$ to the experts based on the task input features $X_t$. Using the current task data $\widetilde{W_i}$, the $i+1$-th branch network $\Phi^{(i)}$ and its corresponding gating network $g_i$ are trained. The gating mechanism ensures that task-specific data is processed by selecting appropriate experts, and the outputs of different experts are dynamically combined through softmax-based weighting:
\begin{align}
\omega_m(X_t, \Theta_t) = \frac{\exp(h_m(X_t, \theta_t^{(m)}))}{\sum_{m'=1}^{M} \exp(h_{m'}(X_t, \theta_t^{(m')}))}
\end{align}

In a given round $t$, a task is randomly selected, and a new expert $m$ is trained using the data of this task. This expert is assigned to integrate the knowledge of all current experts $M$. Each expert participates in task processing based on the output weights from the gating network. The total output of the weighted linear outputs from multiple experts can be expressed as the weighted sum of the outputs of individual experts, fully leveraging the knowledge of multiple experts to produce more stable and accurate decision results. Specifically, in each training round, assuming the linear output of each expert $m$ is $h_m(X_t, \theta_t^{(m)})$ and the contribution of each expert is determined by the weights $\omega_m(X_t, \Theta_t)$. The total output of the weighted linear outputs from multiple experts can be expressed as:
\begin{align}
H(X_t, \Theta_t) = \sum_{m=1}^M \omega_m(X_t, \Theta_t) \cdot h_m(X_t, \theta_t^{(m)})
\end{align}
where $H(X_t, \Theta_t)$ represents the final weighted output, and $\omega_m(X_t, \Theta_t)$ denotes the weight of expert $m$.

In the MoPE model, each task is assigned to an independent expert branch network $\Phi^{(i)}$, whose specific parameters $\theta^{(i)}$ support task-specific optimization. The distributed multi-expert architecture significantly mitigates gradient conflict issues by reducing parameter sharing, ensuring optimization independence among tasks. The core objective of the model is to minimize the total loss across multiple tasks through dynamic task allocation. By leveraging the dynamic gating mechanism, tasks are dynamically assigned to the most relevant expert branches based on task characteristics, effectively reducing parameter interference and gradient conflict risks between unrelated tasks. Additionally, during each training round, the network activates only a few experts relevant to the current task, while keeping the remaining experts frozen. This sparse activation strategy not only improves the computational efficiency of the model but also enhances its adaptability in multi-task environments.

\begin{figure*}[htbp]
\centering
\includegraphics[width=6.8in]{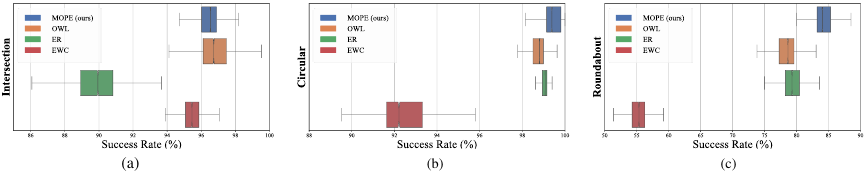}
\caption{The performance of the proposed method is evaluated against baseline models in scenario adaptation, including unprotected left turns (a), sharp circular curves (b), and multi-lane roundabouts (c).}
\label{fig3} 
\end{figure*}

In each training round, the training loss is defined by calculating the difference between the weighted predictions of all experts and the true value $y_t$ under the sample size $n_t$ using the mean squared error (MSE) loss function:
\begin{align}
\mathcal{L}_t(w_t, D_t) = \frac{1}{n_t} \sum_{m \in [M]} \omega_m(X_t, \Theta_t) \| X_t^\top w_t^{(m)} - y_t \|_2^2
\end{align}

Based on this loss formula, the model parameters of each expert are updated according to their weighted contribution to the task outcome. This training loss formula quantifies the model's prediction error on task $t$ by combining the expert weights to compute a weighted sum of the squared Euclidean errors of each expert, normalized to ensure robustness to task scale. The formula integrates expert collaboration and task adaptability, ensuring that the model not only accurately fits the current task during optimization but also enhances overall adaptability. For each expert $m$, the model parameters are updated using the following formula:
\begin{align}
\pi_{t}^{(m)} = \pi_{t-1}^{(m)} + X_t (X_t^\top X_t)^{-1} (y_t - E_t)
\end{align}
where $E_t = \sum_{m' \in [M]} \omega_m(X_t, \Theta_t) X_t^T w_t^{(m')}$ represents the weighted output of all experts. 

The gating network optimizes its parameters $\theta^{(i)}$ using gradient descent, dynamically allocating weights to each expert based on the current task features, which directly affects the total output. By continuously updating the gating network, the model can adaptively optimize task allocation among experts, improving overall prediction accuracy. After completing the training for each task, the model freezes the parameters $\theta^{(0:i)}$ of the $i$-th branch network, preserving its structure and parameters to prevent new tasks from overwriting historical knowledge and avoiding forgetting. Through dynamic task allocation and knowledge sharing, the model effectively balances complex optimization objectives. The parameter update employs a least squares optimization mechanism, combining input features, true outputs, and the weighted predictions of multiple experts to dynamically adjust expert parameters and minimize task errors. The introduction of the pseudo-inverse matrix ensures efficient updates and coordinates knowledge sharing and independence among experts, enabling the model to flexibly adapt to new tasks while retaining existing knowledge, thereby enhancing its adaptability in continual learning environments.

After completing the training for each task, the model freezes the parameters $\theta^{(t)}$ of the expert network for the current task, preventing further optimization. This ensures that the knowledge of historical tasks is not overwritten during the training of subsequent tasks, effectively avoiding the issue of catastrophic forgetting. The total loss after freezing can be decomposed as:
\begin{align}
\mathcal{L} = \sum_{t=1}^T \mathcal{L}_t(\theta^{(t)}) + \sum_{t=1}^T \gamma \|\theta^{(t)} - \theta^{*(t)}\|^2
\end{align}
where $\|\theta^{(t)} - \theta^{*(t)}\|^2$ represents the deviation between the current task parameters and the optimal parameters of historical tasks. $\gamma$ is the regularization weight used to constrain this deviation. After freezing the parameters, $\theta^{(t)}$ is no longer updated, and the second term of the loss function remains zero, thereby effectively preserving the knowledge of historical tasks.

The Mixture of Progressive Experts network facilitates the expansion of new knowledge while effectively retaining prior system knowledge, and the dynamic allocation of modular expert networks enables efficient knowledge transfer and task adaptability in multi-task environments. This architecture is well-suited for real-time model optimization in autonomous driving and adaptation to complex, dynamic scenarios, allowing the system to meet diverse new environmental demands while maintaining safety.


\section{Experiments}

\subsection{Experiments Setup}
The experiments evaluate the Mixture of Progressive Experts in autonomous driving, focusing on scenario adaptation and case generalization. The training environment was constructed utilizing the simulation software MetaDrive \cite{li2022metadrive}. The evaluation scenarios include high-curvature circular, unprotected left-turn intersection, and multi-lane roundabout, each with unique driving characteristics. Surrounding vehicles follow the Intelligent Driver Model (IDM), determining acceleration by distance and relative speed to the front car \cite{kesting2010enhanced}. Based on the dynamic progressive optimization framework in Section II, the experiment employs self-learning experts to generate high-quality data, which is then used to train the continual model designed to adapt over time.

The proposed MoPE model is evaluated against three representative continual learning methods, spanning regularization-based, architecture-based, and replay-based approaches. Elastic Weight Consolidation (EWC) mitigates catastrophic forgetting by penalizing updates to critical weights \cite{riemer2018learning}; OWL employs separate independent policy heads to avoid task conflicts \cite{kessler2022same}; and Experience Replay (ER) stores and replays past data to prevent forgetting during global training \cite{kirkpatrick2017overcoming}.

\subsection{Scenario Adaptation}

Scenario adaptability refers to a model's ability to efficiently respond to environmental changes across diverse driving scenarios, maintaining stable performance even in unseen conditions \cite{niehaus2009scenario}. High-level autonomous driving systems operate in complex, dynamic, and uncertain real-world environments influenced by factors such as road structure, geography, and weather. It demands systems to perform well in known scenarios while demonstrating adaptability for managing unknown or rare situations.
\begin{table}[ht]
\centering
\caption{Comparison of models across various scenarios}
\label{table1}
\renewcommand{\arraystretch}{1.5}  
\begin{tabular}{lcccc}
\toprule
\textbf{Model} & \textbf{Intersection} & \textbf{Circular} & \textbf{Roundabout} \\
\midrule
\textbf{ER} \cite{riemer2018learning} & 89.56\% & 99.01\% & 79.43\% \\
\textbf{EWC} \cite{kirkpatrick2017overcoming} & 95.56\% & 93.04\% & 15.49\% \\
\textbf{OWL} \cite{kessler2022same} & \textbf{96.84\%} & 98.7\% & 55.49\% \\
\rowcolor{gray!10}  
\textbf{MoPE (ours)} & 96.44\% & \textbf{99.43\%} & \textbf{84.17\%} \\
\bottomrule
\end{tabular}
\end{table}
By evaluating the model's performance in diverse scenarios, the advantages of the Mixture of Progressive Experts in scenario adaptability are validated, showcasing its capability to handle dynamic environments with consistent performance. The experiment compares the performance of the proposed MoPE, ER\cite{riemer2018learning}, EWC\cite{kirkpatrick2017overcoming} and OWL\cite{kessler2022same}. The models are trained sequentially on intersections, circulars, and roundabouts using generated RL expert data. Tasks at intersections and circulars are relatively manageable, whereas roundabouts involve more complex interactions between traffic flows.

As can be seen from Fig.\ref{fig3} and Table.\ref{table1}, in the intersection scenario, MoPE sustains stable performance in trained scenarios despite exposure to diverse subsequent datasets, operating on par with OWL and EWC, while ER drops to 90\%. It shows that MoPE mitigates catastrophic forgetting and ensures stability across scenarios. When tested in the circular, all models achieve high success rates, but EWC underperforms due to conflicts in weight updates. Most notably in roundabout, which presents the highest complexity, MoPE attains a 5\% higher success rate. In summary, by leveraging general driving knowledge and dynamic expert weight adaptation, MoPE demonstrates superior adaptability, outperforming others in challenging scenarios while mitigating forgetting.


\subsection{Case Generalization}

Case generalization, the ability to continuously assimilate new data and handle unforeseen corner cases, is essential for ensuring the safety and robustness of autonomous driving systems \cite{jiang2024critical}. Real-world transport environments involve unexpected situations like accidents and extreme weather, often absent from conventional training data. To address these challenges, autonomous systems require efficient learning mechanisms that enable continuous knowledge integration through optimized data acquisition and model refinement. The experiment evaluates the case generalization capability of MoPE by employing the dynamic progressive optimization framework over three training iterations \cite{sun2024generalizing}. The experimental environment simulates urban traffic.The system adapts through failure cases by progressively expanding expert networks, enabling rapid adaptation to corner cases \cite{altekin2012task}. 
\begin{figure}[htbp]
\centering
\includegraphics[width=3.2in]{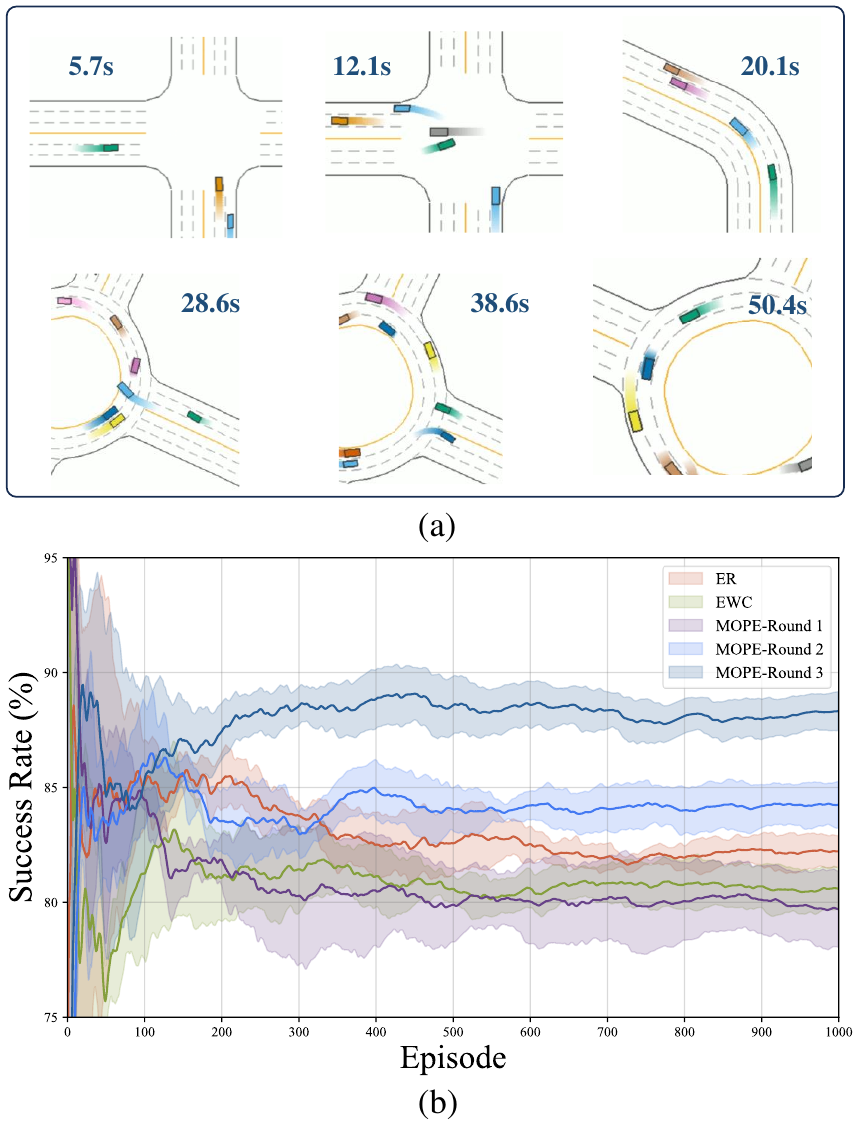}
\caption{Results of the proposed algorithm in case generalization. (a) Dynamic traffic flow featuring typical scenarios described in Section IV.B, each labeled with timestamps. The green rectangle is the ego vehicle. (b) Comparison of the proposed method's performance with baseline models.}
\label{fig4} 
\end{figure}

Over three training iterations, the MoPE model progressively optimizes its handling of failure and corner cases. In the first iteration, it learns basic scenarios but exhibits limitations in complex corner cases due to insufficient data. In subsequent iterations, the expert network expands, improving performance in challenging cases and significantly boosting adaptability. By the third iteration, the system overcomes several previous failures and achieves robust performance.

The MoPE model demonstrates progressive performance improvement across training iterations, as evidenced by the results depicted in Fig.\ref{fig4}. The third iteration achieves the highest success rate of 87\%, indicating that multiple optimization cycles significantly enhance the model's adaptability in complex scenarios. The second iteration reaches 84\%, showing significant progress, while the first iteration exhibits lower success rates and greater fluctuations. The decreasing error bar ranges across iterations indicate MoPE's improved stability. This improvement pattern contrasts sharply with ER and EWC's performance, illustrating limited adaptability to scenario variations. The baseline model performs worse, limited by inadequate identification mechanisms and a deficiency in sustainable data processing. The model shows poor adaptability to new failure cases, particularly in complex driving environments. In contrast, the MoPE model demonstrates evolutionary properties through network expansion and dynamic optimization. Its performance improves continuously, with the third iteration achieving a high success rate, outperforming behavior cloning in adaptability and stability. MoPE efficiently handles failure cases through iterative optimization, demonstrating superior performance in complex autonomous driving tasks.


\subsection{Ablation Study}

\begin{figure}[htbp]
\centering
\includegraphics[width=3.2in]{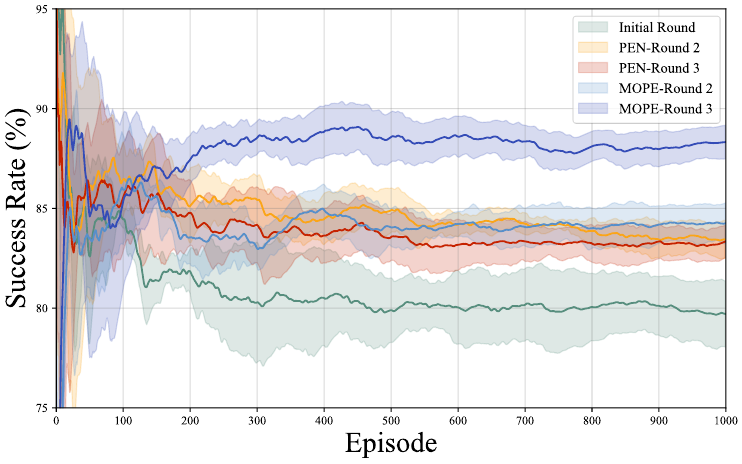}
\caption{The ablation result shows MoPE surpasses PEN, which plateaus without the mixture of experts integration.}
\label{fig5}
\end{figure}

An ablation study was conducted to compare the performance of our proposed MoPE model with the Progressive Expert Network (PEN), which excludes the gating mechanism, highlighting their gaps in subsequent iterations. Without the gating network, PEN struggles with stagnating performance in later iterations, as experts fail to coordinate effectively, as indicated in Fig.\ref{fig5}. In stark contrast, the MoPE model dynamically scales and activates expert modules, progressively enhancing its performance through iterative optimization. The ablation study highlights that the mixture of experts architecture significantly enhances adaptation efficiency, especially in dynamic and complex autonomous driving.


\subsection{Discussion}
\begin{figure}[htbp]
\centering
\includegraphics[width=3.2in]{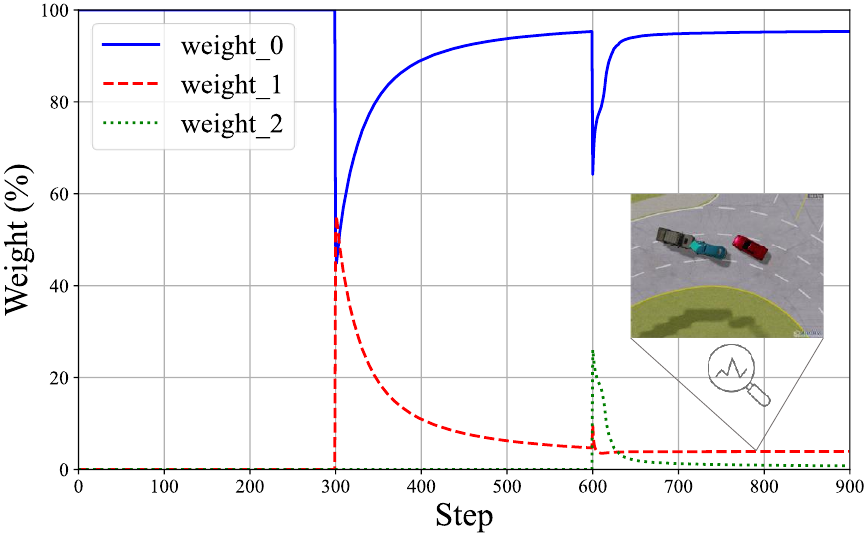} 
\caption{Automatic adjustment of expert weights.}
\label{fig6} 
\end{figure}

Proceeding from the previous analysis, the dynamic variations in expert weights of the MoPE model and their roles will be discussed in this section. Fig.\ref{fig6} shows the dynamic weight changes of each expert agent across different phases, highlighting their respective contributions across diverse cases. The model adopts a phased approach to introduce expert agents, constructing a multi-level decision-making framework that evolves as the training progresses.

The progressive introduction of expert networks employed in this study facilitates adaptability. In the initial stage, only expert agent 0 (with weight $w_0 = 1$) constructs the baseline decision surface for primitive scenes. Subsequently, expert agent 1 ($w_1 \approx 2\%$) is introduced in later stages to focus on rare but high-risk samples. This is achieved while maintaining the stability of the primary expert through a conservative selection strategy of the gating network. Furthermore, expert agent 2 ($w_2 \approx 0.2\%$) is deployed to realize a dynamic sparse activation to address extremely complex cases. It enables on-demand activation for long-tailed scenarios by employing entropy-constrained gating functions, while ensuring the dominance of the base expert \cite{wang2024sparse}. Specifically, the additional agents provide distinct advantages in low-likelihood, high-cost, hazardous scenarios, as their decision sparsity allows for more targeted decision-making.

The gating network employs a weighted selection strategy, favoring trained experts by assigning them higher weights $w_i(x)$, thereby mitigating the risk of erratic decision-making. The model's efficiency is enhanced by the sparse activation mechanism of the gating network, which quantifies the uncertainty of selection probabilities using information entropy $\mathcal{H}(w(x))$:
\begin{equation}
\begin{aligned}
\mathcal{H}(w(x)) = - \sum_{i=1}^{n} w(x)_i \log w(x)_i
\end{aligned}
\end{equation}

Lower information entropy results in higher sparsity, as only a few experts are assigned higher selection probabilities. To address this optimization problem, the Lagrange multiplier method can be applied, incorporating the constraint $\sum_{i=1}^{n} w(x)_i = 1$, leading to the following formulation:
\begin{equation}
\begin{aligned}
\mathcal{J} = \sum_{i=1}^{n} w(x)_i \log w(x)_i + \lambda \left( 1 - \sum_{i=1}^{n} w(x)_i \right)
\end{aligned}
\end{equation}

Through this optimization approach, the model achieves higher sparsity, lower information entropy, and enhanced decision robustness.
\begin{figure}[htbp]
\centering
\includegraphics[width=3.2in]{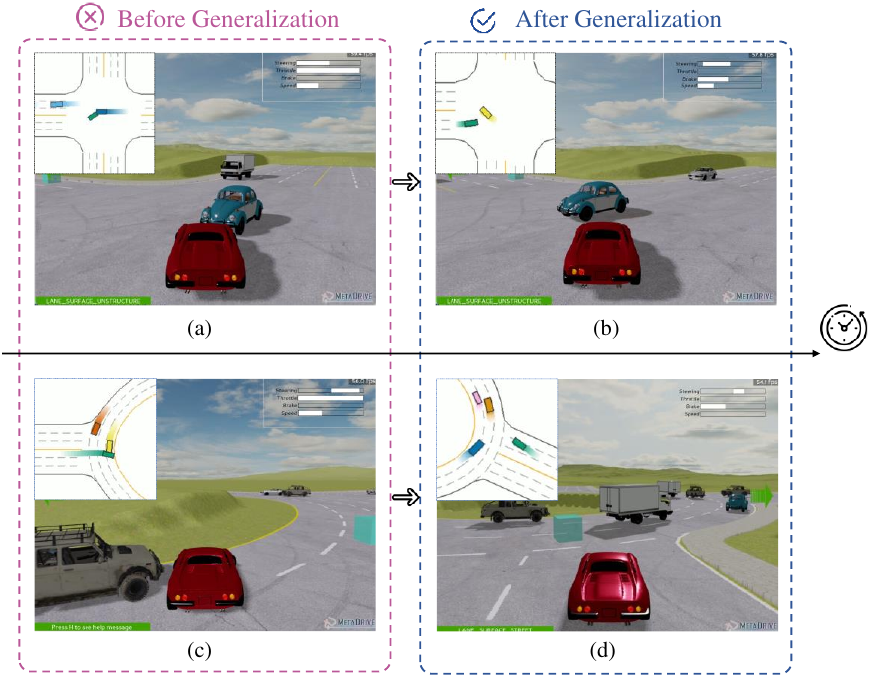}
\caption{Through iterative optimization, MoPE enhances generalization for safety-critical failure cases like intersection conflicts and roundabout merging, with performance validated through simulated renderings and bird’s eye view images.}
\label{fig7} 
\end{figure}

The study summarizes and analyzes two types of typical failure cases, as displayed in Fig.\ref{fig7}, demonstrating how MoPE adapts to diverse driving scenarios by activating sparse experts both before and after generalization. Specifically, (a) and (b) depict conflicts with opposing high-speed vehicles during unprotected left turns at intersections, while (c) and (d) present scenarios of merging into a traffic circle under high traffic density. Prior to generalization, the MoPE model exhibited inefficiency in addressing these challenges. However, post-generalization, the model effectively resolves decision-making issues in such complex and sparse scenarios by leveraging sparse experts to execute critical actions, such as emergency obstacle avoidance and turning maneuvers.

\section{Conclusion}
This paper presents the Mixture of Progressive Experts model as a solution to the adaptation problem in autonomous driving. By leveraging progressively expanding expert network branches and dynamic task weight allocation, MoPE enhances the system's adaptability and efficiency, particularly in complex, dynamic environments. The experimental results clearly indicate that the MoPE model outperforms behavior cloning models, exhibiting a substantial 7.8\% enhancement in decision-making across diverse traffic scenarios. Ultimately, MoPE paves the way for autonomous vehicles to efficiently navigate unknown and complex scenarios, ensuring sustained adaptability and accelerating knowledge transfer across varying environments.

\ifCLASSOPTIONcaptionsoff
  \newpage
\fi

\bibliographystyle{IEEEtran}
\bibliography{reference.bib}



\begin{IEEEbiography}
[{\includegraphics[width=1in,height=1.25in,clip,keepaspectratio]{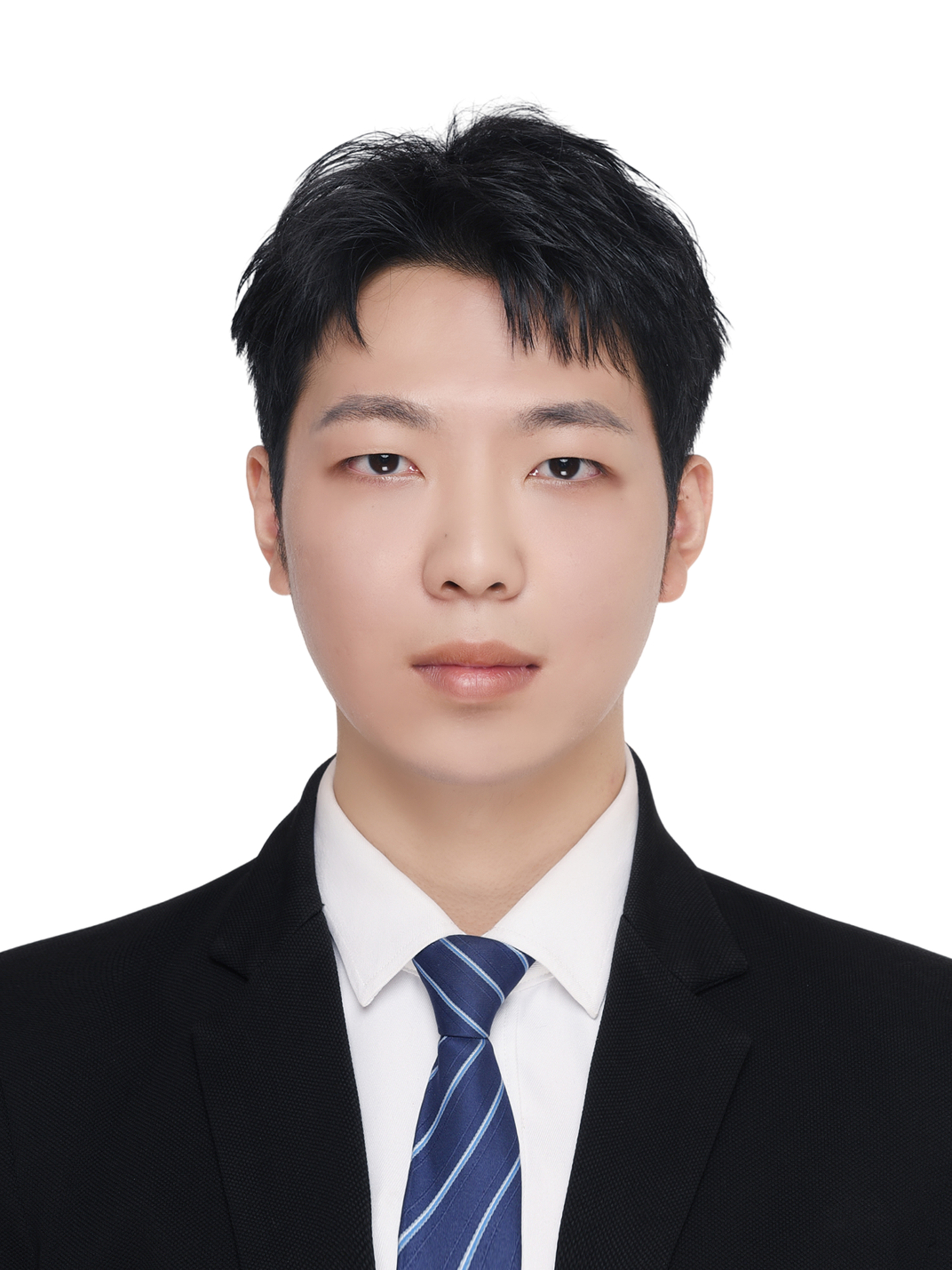}}]{Yixin Cui}
received the B.S. degree from the College of Transportation, Jilin University, Changchun, China, in 2024. He is currently pursuing the Ph.D. degree with the School of Automotive Studies, Tongji University, Shanghai, China. His research interests include reinforcement learning, decision control, intelligent transportation systems, and autonomous vehicles.
\end{IEEEbiography}
\begin{IEEEbiography}
[{\includegraphics[width=1in,height=1.25in,clip,keepaspectratio]{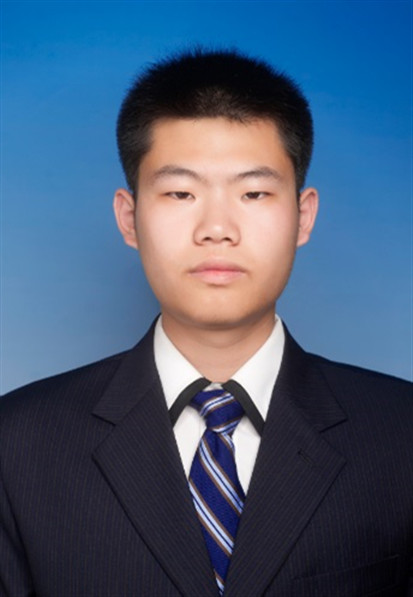}}]{Shuo Yang}
received the B.S. and M.S. degrees (cum laude) from the College of Automotive Engineering, Jilin University, Changchun, China, in 2017. He is currently pursuing the Ph.D. degree with the School of Automotive Studies, Tongji University, Shanghai, China. His research interests include reinforcement learning, autonomous vehicles, intelligent transportation systems, and vehicle dynamics.
\end{IEEEbiography}
\begin{IEEEbiography}
[{\includegraphics[width=1in,height=1.25in,clip,keepaspectratio]{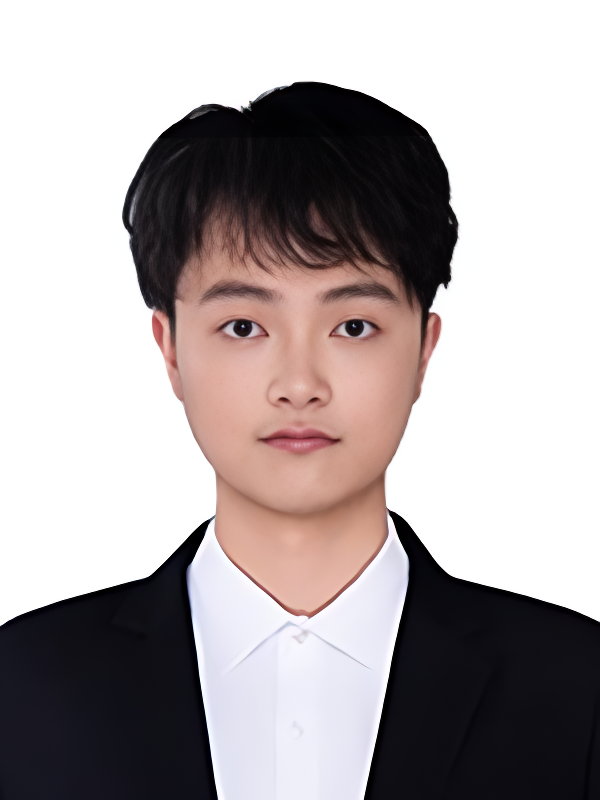}}]{Chi Wan}
is currently pursuing the B.S. degree with the College of Mechanical and Vehicle Engineering, Chongqing University, Chongqing, China. His research interests include machine learning, decision control, intelligent transportation systems, and autonomous vehicles.
\end{IEEEbiography}
\begin{IEEEbiography}
[{\includegraphics[width=1in,height=1.25in,clip,keepaspectratio]{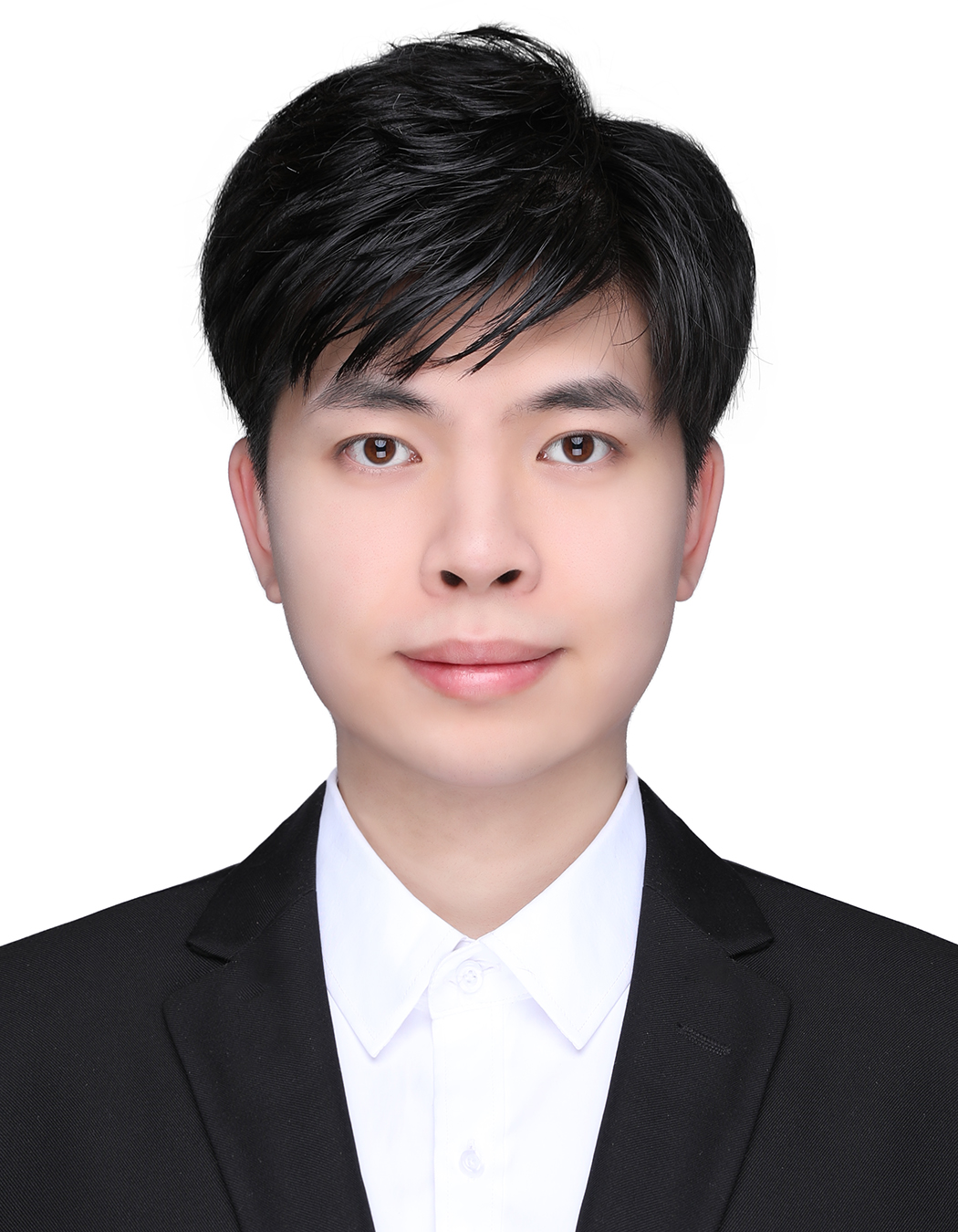}}]{Xincheng Li}
received the B.S. degree in industrial design (vehicle body engineering) from Jilin University, Changchun, China, in 2021. He is currently pursuing the Ph.D. degree with the School of Automotive Studies, Tongji University, Shanghai, China. His research interests include machine learning, reinforcement learning, intelligent transportation systems, and autonomous vehicles.
\end{IEEEbiography}
\begin{IEEEbiography}
[{\includegraphics[width=1in,height=1.25in,clip,keepaspectratio]{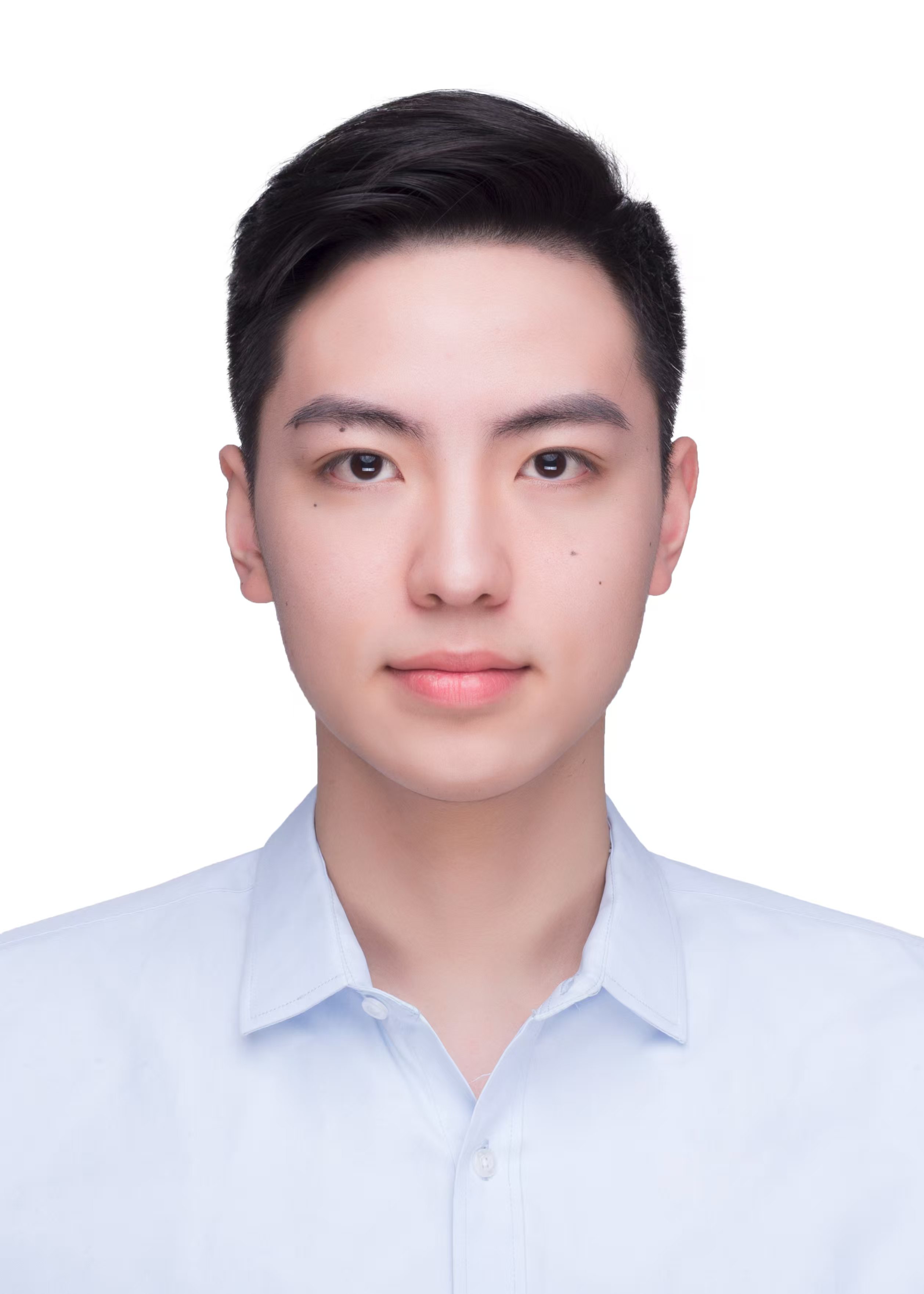}}]{Jiaming Xing}
received the M.S. degree in automotive engineering from Jilin University, Changchun, China, in 2022. He is currently pursuing the Ph.D. degree with the School of Automotive Studies, Tongji University, Shanghai, China. His research interests include continual learning, reinforcement learning, intelligent transportation systems, and autonomous driving.
\end{IEEEbiography}
\begin{IEEEbiography}
[{\includegraphics[width=1in,height=1.25in,clip,keepaspectratio]{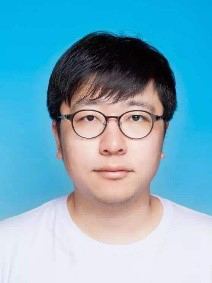}}]{Yuanjian Zhang}
(Member, IEEE) received the M.S. degree in automotive engineering from Coventry University, Coventry, U.K., in 2013, and the Ph.D. degree in automotive engineering from Jilin University, Changchun, China, in 2018. Then, he joined the University of Surrey, Guildford, U.K., as a Research Fellow in advanced vehicle control. He was a Research Fellow at the Sir William Wright Technology Centre, Queen’s University Belfast, Belfast, U.K., and later as a Lecturer (Assistant Professor) in the Department of Aeronautical and Automotive Engineering at Loughborough University, Loughborough, U.K. He is currently a Professor with the School of Automotive Studies, Tongji University, Shanghai, China. He has authored several books and over 50 peer-reviewed journal articles and conference proceedings. His current research interests include advanced control on electric vehicle powertrains, vehicle-environment-driver cooperative control, vehicle dynamic control, and intelligent control for driving assistance systems. 
\end{IEEEbiography}
\begin{IEEEbiography}
[{\includegraphics[width=1in,height=1.25in,clip,keepaspectratio]{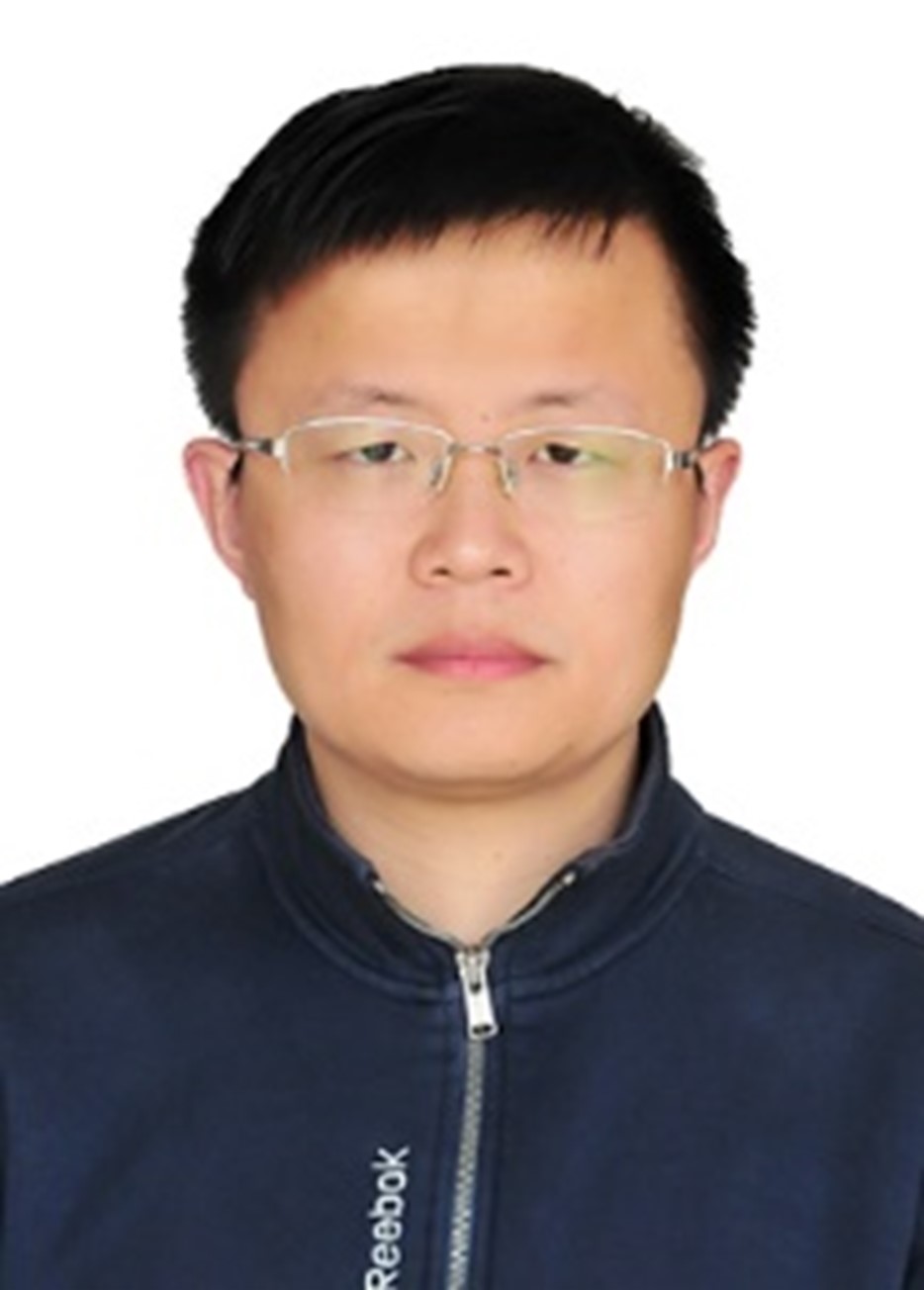}}]{Yanjun Huang}
(Member, IEEE) received the Ph.D. degree from the Department of Mechanical and Mechatronics Engineering, University of Waterloo, Waterloo, ON, Canada, in 2016. He is currently a Professor with the School of Automotive Studies, Tongji University, Shanghai, China. He has published several books and over 80 papers in journals and conferences. His research interests include autonomous driving and artificial intelligence, particularly in decision-making and planning, motion control, and human–machine cooperative driving. He was a recipient of the IEEE Vehicular Technology Society 2019 Best Land Transportation Paper Award. He serves as an Associate Editor for \textit{IEEE Transactions on Intelligent Transportation Systems}, \textit{Proceedings of The Institution of Mechanical Engineers—Part D: Journal of Automobile Engineering}, \textit{IET Intelligent Transport Systems}, \textit{International Journal of Computer Vision (SAE)}, and the Springer book series on connected and autonomous vehicles.
\end{IEEEbiography}

\begin{IEEEbiography}
[{\includegraphics[width=1in,height=1.25in,clip,keepaspectratio]{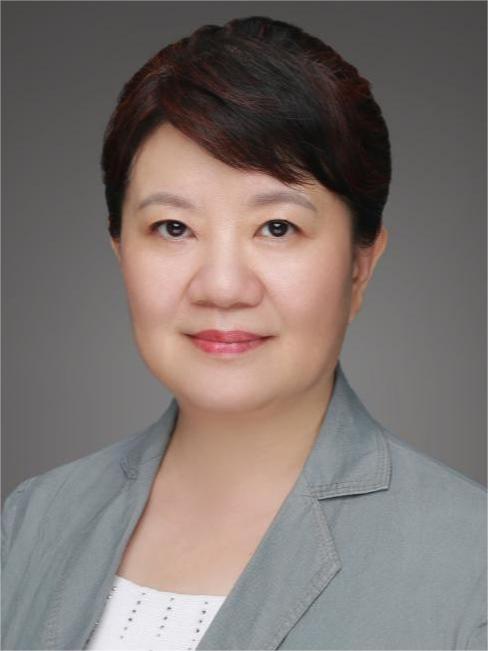}}]{Hong Chen}
(Fellow, IEEE) received the B.S. and M.S. degrees in process control from Zhejiang
University, Hangzhou, China, in 1983 and 1986, respectively, and the Ph.D. degree in system dynamics and control engineering from the University of Stuttgart, Stuttgart, Germany, in 1997. In 1986, she joined Jilin University of Technology, Changchun, China. From 1993 to 1997, she was a Wissenschaftlicher Mitarbeiter with the Institut für Systemdynamik und Regelungstechnik, University of Stuttgart. Since 1999, she has been a Professor at Jilin University, Changchun, and hereafter a Tang Aoqing Professor. Recently, she joined Tongji University as a Distinguished Professor. Her current research interests include model predictive control, nonlinear control, artificial intelligence, and applications in mechatronic systems, e.g., automotive systems.
\end{IEEEbiography}

\end{document}